\documentclass[sigconf,nonacm]{acmart}
\usepackage{multirow}
\usepackage{booktabs}
\usepackage{subcaption}
\usepackage{graphicx}

\renewcommand\footnotetextcopyrightpermission[1]{}
\AtBeginDocument{%
  }

\begin{document}

\title{QDEvo: A Multi-Objective Quality-Diversity Framework for Automated Heuristic Design}

\author{Do Khanh Nam}
\affiliation{%
  \institution{Hanoi University of Science and Technology}
  \city{Hanoi}
  \country{Vietnam}
}
\email{nam.dk225988@sis.hust.edu.vn}

\author{Nguyen Tran Minh Nhat}
\affiliation{%
  \institution{Hanoi University of Science and Technology}
  \city{Hanoi}
  \country{Vietnam}
}
\email{nhat.ntm235986@sis.hust.edu.vn}

\author{Pham Vu Tuan Dat}
\affiliation{%
  \institution{Hanoi University of Science and Technology}
  \city{Hanoi}
  \country{Vietnam}
}
\email{dat.pvt251081m@sis.hust.edu.vn}

\author{Long Doan}
\affiliation{%
  \institution{George Mason University}
  \city{Fairfax}
  \state{Virginia}
  \country{USA}
}
\email{ldoan5@gmu.edu}

\author{Huynh Thi Thanh Binh}
\affiliation{%
  \institution{Hanoi University of Science and Technology}
  \city{Hanoi}
  \country{Vietnam}
}
\email{binhht@soict.hust.edu.vn}

\renewcommand{\shortauthors}{Do Khanh Nam et al.}
\begin{abstract}
    The integration of Large Language Models (LLMs) with evolutionary computation has emerged as a powerful paradigm for automated heuristic design in combinatorial optimization.
    However, existing approaches suffer from mode collapse, converging to homogeneous populations that lack semantic diversity and fail to explore the full algorithmic space.
    We propose Quality-Diversity Evolution (QDEvo), a multi-objective framework that integrates Quality-Diversity optimization with LLM-driven heuristic search, maintaining an unbounded archive of semantically diverse algorithms using pre-trained code embeddings and incorporating hierarchical self-reflection to guide the evolutionary process.
    Extensive experiments across standard benchmarks and real-world industrial applications demonstrate that QDEvo significantly outperforms state-of-the-art methods in both Hypervolume and Inverted Generational Distance metrics.
    Our framework enables the discovery of heuristics that are simultaneously high-performing, computationally efficient, and semantically diverse, providing practitioners with a rich portfolio of solutions for complex optimization problems.
\end{abstract}

\ccsdesc[500]{Computing methodologies~Heuristic function construction}
\ccsdesc[500]{Theory of computation~Discrete optimization}

\keywords{Large Language Models, Quality-Diversity Optimization, Automated Heuristic Design, Evolutionary Computation, Multi-objective Optimization, Reflective Learning}

\maketitle

\section{Introduction}
Large Language Models (LLMs) have demonstrated remarkable capabilities across diverse tasks~\cite{zhao2023survey}, with their code generation proficiency particularly advancing Automated Heuristic Design (AHD). By combining LLMs with Evolutionary Computation (EC), a ``Thought-to-Code'' paradigm has emerged where natural language reasoning directly guides the generation of human-readable, executable heuristics to solve combinatorial optimization problems (COPs). Starting with FunSearch~\cite{romera2024mathematical}, subsequent works~\cite{liuevolution, ye2024reevo, dat2025hsevo, zheng2025monte} have advanced this direction, though these methods initially focused on single-objective settings, optimizing only solution quality.

Recent efforts have begun extending LLM-based AHD to multi-objective settings, simultaneously optimizing solution quality and computational efficiency~\cite{yao2025multi,kiet2026motif} or multi-objective solutions~\cite{hieu2026pareto}. However, the exploration-exploitation dilemma remains a fundamental challenge across both single- and multi-objective approaches. These methods tend to converge to a narrow set of high-performing but homogeneous solutions, as their selection mechanisms operate solely in the objective space without considering the underlying algorithmic structure. This lack of semantic diversity severely constrains the search process, preventing the discovery of novel algorithmic paradigms and causing premature convergence to local optima.

To address this, we propose Quality-Diversity Evolution (QDEvo), a novel framework that integrates Quality-Diversity optimization principles~\cite{pugh2016quality} with LLM-driven heuristic search. QDEvo maintains an unbounded archive of algorithms that are both high-performing and semantically distinct, using pre-trained code embeddings to measure algorithmic similarity and actively enforce diversity. This provides practitioners with diverse algorithmic strategies and enables the discovery of ``stepping stone'' solutions that lead to superior heuristics through further evolution. The framework further incorporates Hierarchical Reflective Memory to learn from both successes and failures across the evolutionary process. Our contributions are as follows:
\begin{itemize}
    \item \textbf{Quality-Diversity Evolution Framework:} We propose the first LLM-based heuristic generation framework that leverages semantic code embeddings to enforce population diversity, preventing mode collapse and enabling broader exploration of the algorithmic space.
    \item \textbf{Multi-Objective Optimization:} We also demonstrate that QDEvo can simultaneously optimize for both single-objective solution quality and computational efficiency, or for inherently multi-objective problems.
    \item \textbf{Hierarchical Reflective Memory:} A structured memory architecture that enables the LLM to learn from successful strategies, synthesize long-term experiences, and explicitly avoid previously encountered errors.
\end{itemize}

\begin{figure*}[t]
    \centering
    \includegraphics[width=\linewidth]{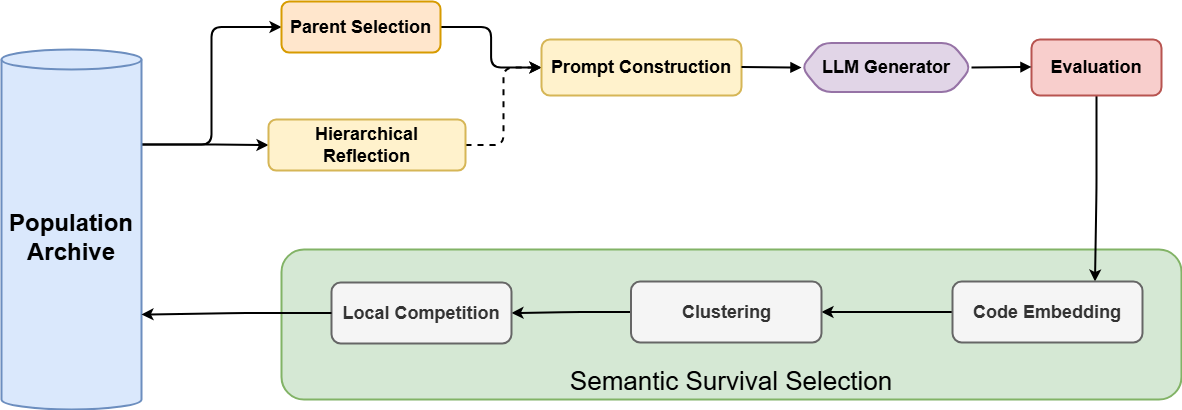}
    \caption{Overview of the QDEvo framework. The evolutionary loop begins with \textbf{Parent Selection} from the Population Archive. Selected parents, along with insights from \textbf{Hierarchical Reflection}, feed into \textbf{Prompt Construction} (dashed arrow indicates the feedback loop). The \textbf{LLM Generator} produces new candidate heuristics, which undergo \textbf{Evaluation}. Finally, \textbf{Semantic Survival Selection} (comprising Code Embedding, Clustering, and Local Competition) determines which individuals return to the archive for subsequent generations.}
    \label{fig:qdevo}
\end{figure*}

\section{Methodology}
\subsection{Evolutionary Process}

The QDEvo framework operates as an iterative evolutionary loop, as illustrated in Figure~\ref{fig:qdevo}. Each individual is represented as a \textit{Thought-to-Code} pair $(T, C)$, where $T$ denotes the algorithmic thought a natural language description of the heuristic strategy and $C$ represents the executable Python code implementing that strategy~\cite{liuevolution}. This dual representation enables the LLM to reason about algorithm design at both conceptual and implementation levels. Each individual is further augmented with an embedding vector for diversity-based selection and a score vector $[f_{\text{quality}}, f_{\text{runtime}}]$ for multi-objective fitness evaluation.

The evolutionary process begins with the \textbf{Population Archive}, an unbounded structure that stores all non-dominated individuals discovered throughout the search. Unlike fixed-size populations in previous works, this archive naturally grows to accommodate diverse solutions while pruning dominated individuals during survival selection. Following standard practice in LLM-based evolutionary methods \cite{liuevolution,romera2024mathematical}, the archive is bootstrapped via zero-shot generation with adaptive sampling, iteratively generating heuristics from only the task description until a minimum number of valid individuals pass evaluation. A higher temperature ($\tau = 1.0$) is used during initialization to maximize diversity.

From this archive, the \textbf{Parent Selection} module samples individuals for variation. For mutation, a single parent is selected uniformly at random from valid individuals, promoting exploration across the behavioral space. For crossover, two parents are randomly sampled and ordered based on Pareto dominance:
\begin{equation}
(P_{\text{better}}, P_{\text{worse}}) = 
\begin{cases}
(P_1, P_2) & \text{if } P_1 \succ P_2 \\
(P_2, P_1) & \text{otherwise}
\end{cases}
\end{equation}
This ordering enables the LLM to identify superior design patterns from the better parent while potentially rescuing beneficial components from the worse parent.

The selected parents then flow into \textbf{Prompt Construction}, where task context, parent information, and reflection insights are assembled. Critically, insights from the \textbf{Hierarchical Reflective Memory} (detailed in Section~\ref{sec:reflection_memory}) feed into this stage via a feedback loop, enabling the LLM to leverage accumulated knowledge from prior generations.

The constructed prompts are passed to the \textbf{LLM Generator}, which serves as the variation operator. QDEvo employs one crossover and three mutation strategies: \textit{Crossover} combines design elements from both parents guided by reflection insights; \textit{Random Mutation} encourages novelty through radical structural changes; \textit{Reflection-Guided Mutation} leverages accumulated experience to guide modifications; and \textit{Runtime Mutation} focuses exclusively on computational efficiency through vectorization and caching optimizations.

Each generated heuristic then undergoes \textbf{Evaluation} with bi-objective assessment:
\begin{equation}
\mathbf{f}(C) = [f_{\text{quality}}(C), f_{\text{runtime}}(C)]
\end{equation}
where $f_{\text{quality}}$ measures effectiveness on the target optimization problem and $f_{\text{runtime}}$ measures computational efficiency as the negative of execution time. Invalid individuals (syntax errors, runtime exceptions, timeouts) receive $f = [-\infty, -\infty]$, with their error information captured for the error reflection mechanism.

Finally, evaluated individuals enter the \textbf{Semantic Survival Selection} stage (detailed in Section~\ref{sec:sqd_selection}). This module transforms each heuristic into a semantic embedding, clusters similar solutions together, and applies local Pareto competition within each cluster. Surviving non-dominated individuals return to the Population Archive, completing the evolutionary cycle and enabling continuous discovery of diverse, high-quality heuristics.

\subsection{Hierarchical Reflective Memory}
\label{sec:reflection_memory}

As demonstrated in~\cite{ye2024reevo}, verbal gradients play an important role in LLM-based AHD. QDEvo extends this concept by incorporating a three-level hierarchical reflection mechanism that enables progressive learning from evolutionary history. The reflection outputs feed into prompt construction via a feedback loop, enabling the LLM to leverage accumulated knowledge for improved generation.

\noindent \textbf{Flash Reflection} provides immediate feedback at each generation. A diverse sample of individuals spanning the performance spectrum is analyzed, where the LLM compares pairs of better versus worse heuristics, identifying patterns that distinguish high-performing from low-performing designs. The output consists of detailed comparison insights (e.g., ``Comparing best vs worst, we observe...'') and concise actionable advice.

\noindent \textbf{Comprehensive Reflection} aggregates experiences from multiple generations, synthesizing them into structured guidance with four components: \textit{Keywords} (key concepts), \textit{Advice} (recommended strategies), \textit{Avoid} (pitfalls to prevent), and \textit{Explanation} (reasoning behind the guidance). This reflection distinguishes between experiences that led to Pareto front improvements versus those that did not, enabling long-term learning of successful design principles.

\noindent \textbf{Error Reflection} tracks failed individuals with their error information (error type, message, traceback). A repair mechanism attempts to fix failed heuristics using error-aware prompts, distilling runtime failures into explicit avoidance rules. This prevents the LLM from repeating common mistakes and systematically guides future generations toward valid solutions.

\subsection{Semantic Survival Selection}
\label{sec:sqd_selection}

Each heuristic is canonicalized (removing comments, normalizing formatting via AST parsing) and embedded into a dense vector $\mathbf{v}$ using a pre-trained code embedding model. Individuals are partitioned into semantic clusters based on cosine similarity:
\begin{equation}
    \text{sim}(P_i, P_j) = \frac{\mathbf{v}_i \cdot \mathbf{v}_j}{\|\mathbf{v}_i\| \|\mathbf{v}_j\|}
\end{equation}
An individual joins cluster $C_k$ only if its similarity to \textbf{all} members exceeds threshold $\delta{=}0.95$; otherwise a new cluster is created:
\begin{equation}
    P_{new} \in C_k \iff \forall P_j \in C_k, \ \text{sim}(P_{new}, P_j) \ge \delta
\end{equation}
Within each cluster, only Pareto non-dominated individuals survive, ensuring that semantically similar solutions compete locally while diversity is preserved across algorithmic niches~\cite{pugh2016quality}.

\section{Experiments}

We compare QDEvo against MEoH~\cite{yao2025multi} (state-of-the-art LLM-based method) and classic evolutionary algorithms (NSGA-II, MOEA/D) across three benchmark categories: (1)~\textit{single-objective COPs} reformulated as bi-objective (quality vs.\ runtime), including TSP (constructive and GLS), CVRP, and Online Bin Packing; (2)~\textit{multi-objective COPs} including Bi-TSP and Bi-KP; and (3)~\textit{real-world tasks} including Airline Crew Pairing and Intra-operator Parallelism\footnote{ASPLOS 2025 Programming Contest: Intra-Operator Parallelism Optimization. \url{https://sites.google.com/view/asplos25contest}. Accessed 2025.}. Performance is measured by Hypervolume (HV, higher is better) and Inverted Generational Distance (IGD, lower is better). All methods use Codestral~2 as the LLM, with 300 max samples and population size 30.

\subsection{Results}

\begin{table*}[ht]
  \centering
  \caption{Overall Performance of QDEvo vs. Baselines across Constructive, Multi-objective, 
  and Real-world Benchmarks. Best results for Hypervolume (HV $\uparrow$) and IGD ($\downarrow$) are 
  highlighted in bold.}
  \label{tab:all_results}
  \resizebox{0.95\textwidth}{!}{%
  \begin{tabular}{l|cc|cc|cc|cc}
  \toprule
  \multirow{2}{*}{\textbf{Problem}} & 
  \multicolumn{2}{c|}{\textbf{MOEA/D}} & \multicolumn{2}{c|}{\textbf{NSGA-II}} & 
  \multicolumn{2}{c|}{\textbf{MEoH}} & \multicolumn{2}{c}{\textbf{QDEvo}} \\
  & HV $\uparrow$ & IGD $\downarrow$ & HV $\uparrow$ & IGD $\downarrow$ & HV $\uparrow$ & IGD $\downarrow$ & HV $\uparrow$ & IGD $\downarrow$ \\
  \midrule
  \multicolumn{9}{c}{\textit{Constructive Heuristics (Quality vs. Efficiency)}} \\
  \midrule
  TSP-C 100 & 1.205$\pm$0.007 & 0.203$\pm$0.067 & 1.206$\pm$0.002 & 0.230$\pm$0.088 & 1.206$\pm$0.001 & 0.252$\pm$0.038 & \textbf{1.207$\pm$0.002} & \textbf{0.187$\pm$0.033} \\
  TSP-C 200 & 1.204$\pm$0.000 & 0.172$\pm$0.019 & 1.205$\pm$0.000 & 0.169$\pm$0.013 & 1.204$\pm$0.001 & 0.168$\pm$0.003 & \textbf{1.206$\pm$0.002} & \textbf{0.134$\pm$0.054} \\
  TSP-GLS 100 & 1.195$\pm$0.003 & 0.453$\pm$0.237 & 1.199$\pm$0.003 & 0.193$\pm$0.069 & 1.196$\pm$0.000 & \textbf{0.155$\pm$0.059} & \textbf{1.208$\pm$0.002} & 0.163$\pm$0.062 \\
  CVRP-C 100 & 1.198$\pm$0.000 & 0.517$\pm$0.056 & 1.197$\pm$0.005 & 0.757$\pm$0.378 & 1.196$\pm$0.002 & 0.757$\pm$0.317 & \textbf{1.204$\pm$0.006} & \textbf{0.480$\pm$0.158} \\
  CVRP-C 200 & 1.190$\pm$0.001 & 0.980$\pm$0.716 & 1.188$\pm$0.001 & 0.883$\pm$0.409 & 1.188$\pm$0.002 & 1.023$\pm$0.647 & \textbf{1.194$\pm$0.011} & \textbf{0.237$\pm$0.004} \\
  OBP 5k & 1.187$\pm$0.003 & 0.727$\pm$0.012 & 1.185$\pm$0.002 & 0.638$\pm$0.062 & 1.188$\pm$0.001 & 0.614$\pm$0.118 & \textbf{1.196$\pm$0.003} & \textbf{0.467$\pm$0.286} \\
  \midrule
  \multicolumn{9}{c}{\textit{Multi-objective Benchmarks}} \\
  \midrule
  Bi-TSP 100 & 1.067$\pm$0.042 & 0.793$\pm$0.304 & 1.104$\pm$0.002 & 0.564$\pm$0.079 & 1.096$\pm$0.034 & 0.577$\pm$0.327 & \textbf{1.147$\pm$0.042} & \textbf{0.400$\pm$0.147} \\
  Bi-TSP 200 & 1.160$\pm$0.006 & 0.280$\pm$0.019 & 1.150$\pm$0.002 & 0.429$\pm$0.024 & 1.162$\pm$0.011 & 0.285$\pm$0.050 & \textbf{1.178$\pm$0.018} & \textbf{0.241$\pm$0.082} \\
  Bi-KP 200 & 1.089$\pm$0.033 & 0.606$\pm$0.385 & 1.068$\pm$0.048 & 0.872$\pm$0.296 & 1.105$\pm$0.011 & 0.514$\pm$0.096 & \textbf{1.130$\pm$0.026} & \textbf{0.397$\pm$0.040} \\
  \midrule
  \multicolumn{9}{c}{\textit{Real-world Applications}} \\
  \midrule
  Crew Pairing & 1.184$\pm$0.026 & 0.277$\pm$0.190 & 1.188$\pm$0.017 & 0.281$\pm$0.077 & \textbf{1.202$\pm$0.012} & 0.269$\pm$0.093 & 1.194$\pm$0.006 & \textbf{0.110$\pm$0.030} \\
  Intra-Op & 1.209$\pm$0.000 & 0.687$\pm$0.121 & 1.209$\pm$0.000 & 0.445$\pm$0.134 & 1.209$\pm$0.000 & 0.345$\pm$0.005 & \textbf{1.210$\pm$0.000} & \textbf{0.314$\pm$0.106} \\
  \bottomrule
  \end{tabular}%
  }
\end{table*}

Table~\ref{tab:all_results} summarizes QDEvo's performance across representative benchmarks. QDEvo achieves the highest HV on the majority of instances across all three categories. For constructive heuristics, QDEvo attains 73.2\% lower IGD than NSGA-II on CVRP-C 200 ($0.237$ vs.\ $0.883$). On multi-objective benchmarks, QDEvo improves HV by 1.4\% over MEoH on Bi-TSP-200 ($1.178$ vs.\ $1.162$) and reduces IGD by 15.4\%. For real-world tasks, QDEvo achieves 59.1\% lower IGD than MEoH on Crew Pairing ($0.110$ vs.\ $0.269$), indicating superior coverage of the trade-off structure.

\subsection{Ablation Study}

We compare QDEvo's semantic embedding (\textit{Embedding}) against two variants: \textit{Cell} (structural code metrics as MAP-Elites descriptors) and \textit{NSGAII} (crowding distance only, no QD archive).

\begin{table}[h]
\centering
\caption{Ablation: Semantic Embedding vs.\ Structural Metrics (Cell) and NSGA-II. Bold = best.}
\label{tab:ablation_study}
\resizebox{\columnwidth}{!}{%
\begin{tabular}{l|l|cc}
\toprule
\textbf{Problem} & \textbf{Algorithm} & \textbf{HV $\uparrow$} & \textbf{IGD $\downarrow$} \\
\midrule
\multirow{3}{*}{TSP-C 200} 
& NSGAII & 1.204$\pm$0.001 & 0.349$\pm$0.031 \\
& Cell & 1.202$\pm$0.003 & 0.373$\pm$0.016 \\
& Embedding & \textbf{1.206$\pm$0.002} & \textbf{0.333$\pm$0.054} \\
\midrule
\multirow{3}{*}{CVRP-C 200} 
& NSGAII & 1.189$\pm$0.005 & 0.328$\pm$0.002 \\
& Cell & 1.190$\pm$0.006 & \textbf{0.287$\pm$0.055} \\
& Embedding & \textbf{1.194$\pm$0.011} & 0.288$\pm$0.081 \\
\midrule
\multirow{3}{*}{TSP-GLS 100} 
& NSGAII & 1.089$\pm$0.000 & 0.092$\pm$0.005 \\
& Cell & 1.089$\pm$0.001 & 0.102$\pm$0.005 \\
& Embedding & \textbf{1.123$\pm$0.047} & \textbf{0.073$\pm$0.045} \\
\midrule
\multirow{3}{*}{OBP 10k} 
& NSGAII & 1.196$\pm$0.005 & \textbf{0.310$\pm$0.017} \\
& Cell & 1.191$\pm$0.003 & 0.363$\pm$0.090 \\
& Embedding & \textbf{1.199$\pm$0.009} & 0.366$\pm$0.076 \\
\bottomrule
\end{tabular}%
}
\end{table}

The semantic embedding approach achieves the highest HV across all instances (Table~\ref{tab:ablation_study}). The gap is most pronounced on TSP-GLS 100, where Embedding attains HV of $1.123$ vs.\ $1.089$ for both alternatives---a 3.1\% improvement reflecting the discovery of fundamentally different algorithmic strategies. Structural code metrics (Cell) fail to distinguish semantically distinct algorithms with similar control flow, while NSGAII maintains spread along the front but cannot expand into unexplored regions.

\section{Conclusion}
In this paper, we introduced QDEvo, a novel multi-objective Quality-Diversity framework for automated heuristic design. By integrating Large Language Models with a semantic archive and hierarchical reflective memory, QDEvo effectively addresses the challenges of mode collapse and single-objective over-optimization found in previous approaches. Our extensive experiments across standard combinatorial benchmarks and complex real-world industrial tasks demonstrate that QDEvo significantly outperforms state-of-the-art baselines, producing heuristics that are not only high-performing but also computationally efficient and semantically diverse. The ablation study further confirms the critical role of semantic embeddings in maintaining a rich population of algorithmic strategies. Future work will explore the application of QDEvo to even broader classes of problems and the integration of more advanced reasoning capabilities into the evolutionary process. The source code for QDEvo is publicly available at \url{https://github.com/datphamvn/QDEvo}.

\bibliographystyle{ACM-Reference-Format}
\bibliography{references_short}

\end{document}